\documentclass[review]{elsarticle}

\usepackage{lineno,hyperref}
\usepackage{multirow}
\usepackage{amsmath,amsfonts}
\usepackage{booktabs}
\usepackage{multirow}
\usepackage{pifont}
\usepackage{bbding}
\usepackage{bbm}
\usepackage{bm}
\usepackage{array}
\usepackage{color}
\usepackage{epsfig}
\usepackage{graphicx}
\usepackage{amsmath}
\usepackage{amssymb}
\usepackage{subfig}
\usepackage[figuresright]{rotating}

\journal{Pattern Recognition}

\begin{document}
\begin{frontmatter}
\title{CPR-Coach: Recognizing Composite Error Actions based on Single-class Training}

\author[add1]{Shunli Wang}
\author[add2]{Qing Yu}
\author[add1]{Shuaibing Wang}
\author[add1]{Dingkang Yang}
\author[add1]{Liuzhen Su}
\author[add1]{Xiao Zhao}
\author[add1]{Haopeng Kuang}
\author[add3]{Peixuan Zhang}
\author[add1]{Peng Zhai}
\author[add1,add3,add4]{Lihua Zhang\corref{mycorrespondingauthor}}
\cortext[mycorrespondingauthor]{Corresponding author.}
\ead{slwang19@fudan.edu.cn, lihuazhang@fudan.edu.cn}

\address[add1]{Academy for Engineering and Technology, Fudan University, Shanghai, 200433, China}
\address[add2]{Zhongshan Hospital Affiliated with Fudan University, Shanghai, 200032, China}
\address[add3]{Changchun Boli Technologies Co., Ltd., Changchun, 130000, China}
\address[add4]{Engineering Research Center of AI and Robotics, Ministry of Education,
Shanghai, 200433, China}

\begin{abstract}
The fine-grained medical action analysis task has received considerable attention from pattern recognition communities recently, but it faces the problems of data and algorithm shortage.
Cardiopulmonary Resuscitation (CPR) is an essential skill in emergency treatment.
Currently, the assessment of CPR skills mainly depends on dummies and trainers, leading to high training costs and low efficiency.
For the first time, this paper constructs a vision-based system to complete error action recognition and skill assessment in CPR.
Specifically, we define 13 types of single-error actions and 74 types of composite error actions during external cardiac compression and then develop a video dataset named CPR-Coach.
By taking the CPR-Coach as a benchmark, this paper thoroughly investigates and compares the performance of existing action recognition models based on different data modalities.
To solve the unavoidable \textit{Single-class Training} \& \textit{Multi-class Testing} problem, we propose a human-cognition-inspired framework named ImagineNet to improve the model's multi-error recognition performance under restricted supervision.
Extensive experiments verify the effectiveness of the framework.
We hope this work could advance research toward fine-grained medical action analysis and skill assessment.
The CPR-Coach dataset and the code of ImagineNet are publicly available on \url{https://shunli-wang.github.io/CPR-Coach/}.
\end{abstract}


\end{frontmatter}
\section{Introduction}
Although many human action recognition algorithms \cite{ActionTransformer, posegraph, PR-1, PR-2, PR-3, TSN2016ECCV, i3d2017CVPR, TSM2019ICCV, slowfast2019ICCV} in daily life scenarios have been proposed, the high professionalism and data shortage seriously hinder the development of fine-grained medical action analysis technology \cite{AQA_Survey, JHU-JIGSAWS-2014}.
This paper takes Cardiopulmonary Resuscitation (CPR) as the research example, which is a critical life-saving technique for cardiac and respiratory arrest.
CPR aims to restore the patient's spontaneous breathing and circulation.
According to the American Heart Association (AHA) \footnote{\url{https://www.heart.org/}}, 87.7\% of cardiac arrest occurs in families and public places. Rescuers must conduct CPR within 4 minutes to improve the survival rate of the patient. High-quality and standard CPR is the core of effective treatment, while improper actions will reduce the treatment effectiveness.
%
Traditional CPR skill assessment usually requires the participation of the examiner and the dummy equipped with force sensors, in which the examiner scores the rescuer's body movements, and the force sensors can evaluate the compression frequency and strength. 
The cost of this hybrid evaluation method is too high to conduct large-scale training system deployment \cite{ZQ2_2013, GIT-IJCARS-2016}.
In this paper, we build an intelligent system that automatically identifies wrong actions in CPR during skill training, thus significantly reducing the assessment cost and improving training efficiency.

\begin{figure}[t] \centering 
    \begin{center}
    \includegraphics[width=0.7\linewidth]{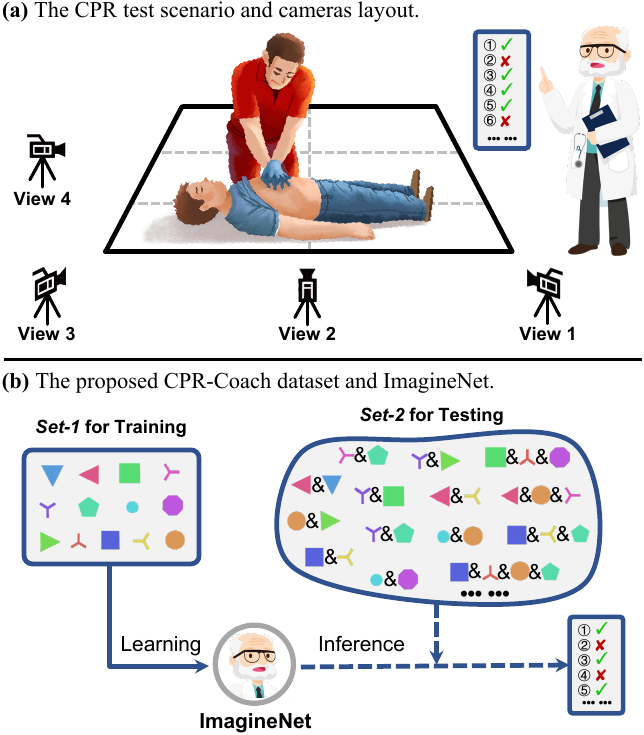}
    \end{center}
    \vspace{-10pt}
	\caption{(a) shows the multi-view capture system. (b) illustrates the structure of the CPR-Coach dataset and the function of the ImagineNet. Each colored mark represents an error action class.}
	\label{Fig1}
\end{figure}

As far as we know, there is no clear definition of specific error types of CPR actions, and no research has been done to explore the vision-based CPR skill assessment.
To fill the research gap, we first identify 13 types of common error actions (shown in Figure \ref{Fig2}(a)) under the guidance of the latest version of \textit{AHA Guidelines for CPR and ECC} \cite{AHA_Guidelines} and professional emergency treatment doctors.
A visual system is constructed to capture videos of the rescue process, as shown in Figure \ref{Fig1}(a).
Based on the above settings, we create a dataset named CPR-Coach, which consists of two parts: \textit{Set-1} that contains single-class actions, and \textit{Set-2} that contains composite error actions.
Figure \ref{Fig1}(b) graphically depicts the structure of the dataset through colored marks.
Note that the square box denotes determined single-class actions in the \textit{Training Set}, while the irregular box denotes the uncertain composite error classes in the \textit{Testing Set}.

Existing action recognition frameworks \cite{ActionTransformer, posegraph, PR-1, PR-2, PR-3, TSN2016ECCV, i3d2017CVPR, TSM2019ICCV, slowfast2019ICCV} have been able to handle the single-class action recognition task.
We can directly migrate these models to CPR-Coach \textit{Set-1} to evaluate the fine-grained errors recognition performance.
However, these models cannot meet the actual application in the CPR test.
In actual CPR skill assessment, rescuers are likely to make multiple mistakes simultaneously, and a qualified coach is supposed to point out all mistakes exactly.
If the number of single errors is 13, the total number of composite errors can reach a frightening 8191 ($\sum_{n=1}^{13} C_{13}^{n} = 2^{13}-1$).
It is impossible to conduct exhaustive data collection to cover all these error combinations. 
In other words, we can't make the labels of the training set consistent with those in test set, such as most classical machine learning tasks.

To solve this dilemma, let us re-think how a real coach works. 
This coach must not have seen all the wrong action combinations, but he can still give the correct judgment according to the single-error action knowledge.
This is because human beings have extremely strong knowledge reasoning and generalization abilities \cite{why}.
Inspired by this, this paper proposes a concise framework named ImagineNet to handle the intractable \textit{Single-class Training} \& \textit{Multi-class Testing} problem properly. 
The function of the ImagineNet is shown in Figure \ref{Fig1}(b).
The essence of the ImagineNet is a human-inspired feature combination training strategy.
As its name implies, it can \textit{Imagine} composite error features based on restricted single-class error actions and achieves high performance in the unseen composite error recognition task.
By regarding \textit{Set-1} as the training set and \textit{Set-2} as the testing set, we can examine the ImagineNet, which plays the role of \textit{Coach}.
Sufficient experimental results confirm the effectiveness of the framework.

The main contributions of this paper are as follows:
\begin{itemize}
\item To the best knowledge, we propose the first dataset named CPR-Coach in the visual CPR assessment task, which supports fine-grained action recognition and composite error recognition tasks.
\item Taking the CPR-Coach dataset as a benchmark, we extensively explore and compare the existing action recognition models based on different modality information.
\item We propose a human-cognition-inspired framework named ImagineNet, which significantly improved the composite error recognition performance under restricted supervision.
\end{itemize}

\section{Related Work}
\subsection{Human Action Recognition}
Video-based Human Action Recognition (HAR) is one of the representative tasks of video understanding.
With the prosperity and development of deep learning methods, more and more action recognition frameworks \cite{ActionTransformer, posegraph, PR-1, PR-2, PR-3, TSN2016ECCV, i3d2017CVPR, TSM2019ICCV, slowfast2019ICCV, TRN2018ECCV, TIN2020AAAI, C3D2015ICCV, timesformer2021ICML, ST-GCN} have been proposed.
Despite the success of previous frameworks on some public HAR benchmarks \cite{ActivityNet, Kinetics400, Sports1M, YouTube-8M, NTU_RGB+D}, the fine-grained recognition performance of these frameworks still remains unexplored \cite{AQA_Survey}, such as in sports and medical fields.
Fortunately, we have now seen the seeds of these specialized studies. 
Benefiting from the availability of sports videos, 
Vakanski \textit{et al.} \cite{UI-PRMD}, Xu \textit{et al.} \cite{FineDiving}, and Shao \textit{et al.} \cite{FineGym} proposed three fine-grained action recognition datasets in sports, respectively. 
In the medical field, some research on surgical workflow recognition and analysis has been proposed \cite{heichole, CATARACTS, Endonet}. These work mainly focus on video analysis of Laparoscopy and Cataract surgeries.
Nevertheless, there are few studies on fine-grained error action recognition in the medical field, 
which can save a lot of resources during the medical skill training and assessment.
To fill this gap, this paper proposes the first dataset named CPR-Coach in CPR skill training and assessment. 
The CPR-Coach dataset contains indistinguishable errors and complex composite error classes, putting forward higher requirements for action recognition models.

\subsection{Action Quality Assessment}
Action Quality Assessment (AQA) aims to identify and score specific skilled actions.
Currently, research on AQA mainly focuse on sports \cite{BPAD, MTL-AQA, FisV-5, AQA-7, UNLV-Dataset, MIT-Diving, TSA-Net} and the medical field \cite{JHU-JIGSAWS-2014, ZQ2_2013, GIT-ISBI-2014, GIT-M2CAI-2016, GIT-IJCARS-2018, JHU-MICCAI-2016, JHU-IJCARS-2015, ASU-ICAMS-2011, ZQ1_2011, ZQ3_2015}. 
Wang \textit{et al.} \cite{AQA_Survey} found that publicly available datasets and algorithms in sports are more than those in the medical field, which is mainly caused by the high professionalism of medical data acquisition.
Existing studies on medical AQA could be divided into three categories: 
surgical skill evaluation \cite{GIT-IJCARS-2016, GIT-ISBI-2014, GIT-M2CAI-2016, GIT-IJCARS-2018, GIT-IJCARS-2017} under the Objective Structured Assessment of Technical Skill (OSATS) system \cite{OSATS},  %
operating skills identification based on \textit{Da Vinci} surgical systems \cite{JHU-JIGSAWS-2014, JHU-MICCAI-2016, JHU-IJCARS-2015, JHU-MIA-2013, JHU-IPCAI-2014}, 
and skill assessment in laparoscopic surgery \cite{ZQ2_2013, ASU-ICAMS-2011, ZQ1_2011, ZQ3_2015, ASU-MMVR-2013, ASU-IUI-2013}.
All these benchmarks are listed in Table \ref{Tab2-Statistics}.
These research only determined the level of Expert/Medium/Novice to rate medical actions and did not conduct detailed analysis.
Users of these systems do not know where they need improvement, so the usage scenarios of these systems are very limited.
This paper proposes the first fine-grained error recognition dataset in CPR and defines the basic form of the composite error action assessment problem.
Note that the CPR test focuses more on specific errors and is not suitable for judging through scores and rated classes. So we extended the concept of AQA to CPR in this work.

\subsection{Multi-Label Learning Algorithms}
Different from traditional classification tasks, multi-label learning faces the challenge of exponential growth in the number of class label spaces \cite{Review-TPAMI-2014, Review-TPAMI-2022}.
Existing solutions are mainly divided into two categories:
Convert the multi-label problem into multiple independent binary classification problems \cite{Multilabel-PR-2004, Multilabel-ICML-2010, Multilabel-SIGKDD-2010}, or improve the algorithm to adapt to multi-label data \cite{Multilabel-TMIS-2011, Multilabel-AAAI-2010, Multilabel-CVPR-2006}.
Although the task of \textit{Single-class Training} \& \textit{Multi-class Testing}  to be solved in this paper is also a multi label classification problem, there is only one single class sample in the training set, so it puts forward higher requirements for the model.
The proposed ImagineNet follows an algorithm transformation strategy and thoroughly improves the recognition performance through feature-combining strategies.

\section{CPR-Coach Dataset}
\begin{figure}
    \begin{center}
    	\includegraphics[width=\linewidth]{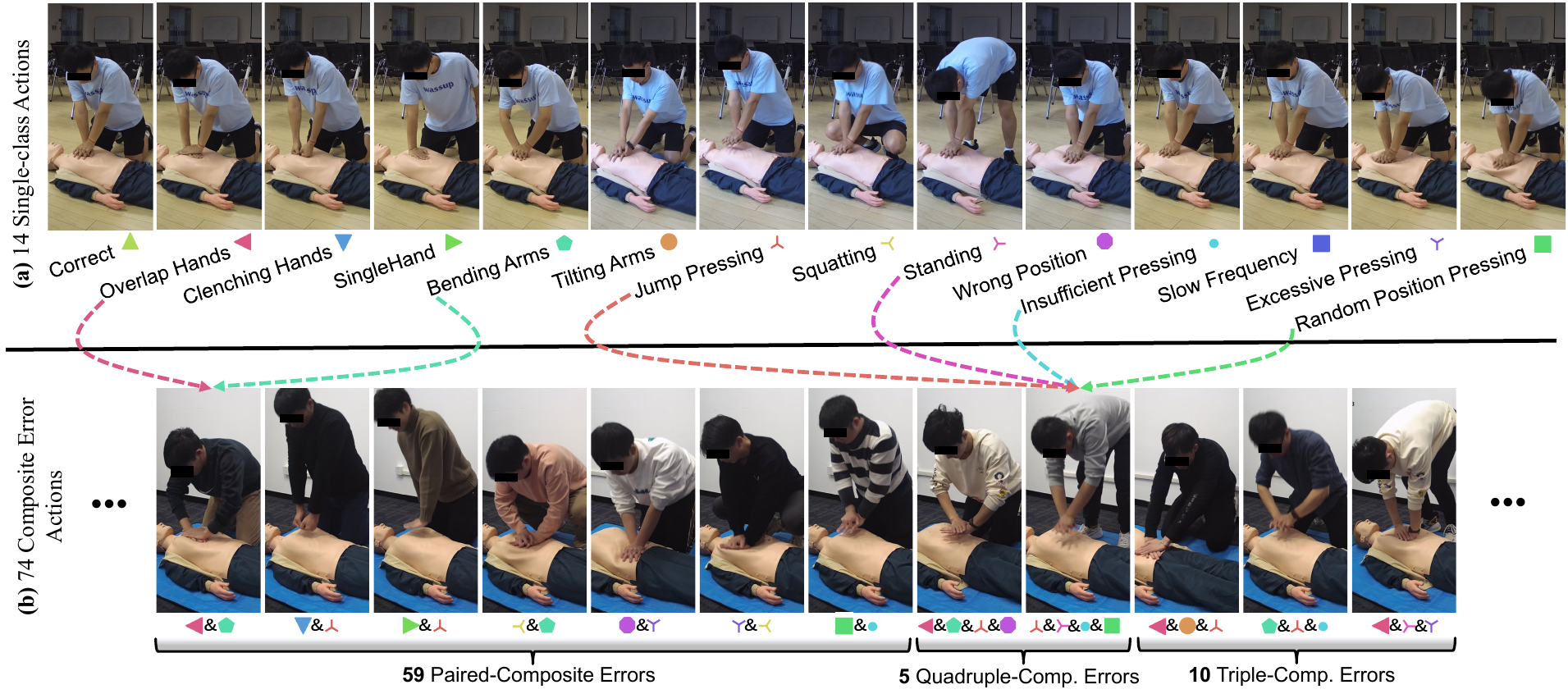}
    \end{center}
	\caption{Structure of the CPR-Coach.
	(a) \textit{Set-1} consists of a \textit{Correct} class and 13 types of single-error actions. (b) \textit{Set-2} consists of 74 composite error actions (59 paired-, 10 triple-, and 5 quadruple-composite errors). 
	For clarity, different marks with different colors are adopted to represent 14 single classes.  
    This marking method is the same elsewhere.
	Due to space limitations, this figure only shows the generation process of one paired- and one quadruple-composite error actions.
	}
	\label{Fig2}
\end{figure}

As shown in Figure \ref{Fig2}, the proposed CPR-Coach dataset is divided into two parts: \textit{Set-1} that contains 1 type of correct action and 13 types of single-error actions, and \textit{Set-2} that contains 74 types of composite error actions.

\begin{figure}[t] \centering
	\includegraphics[width=1.0\linewidth]{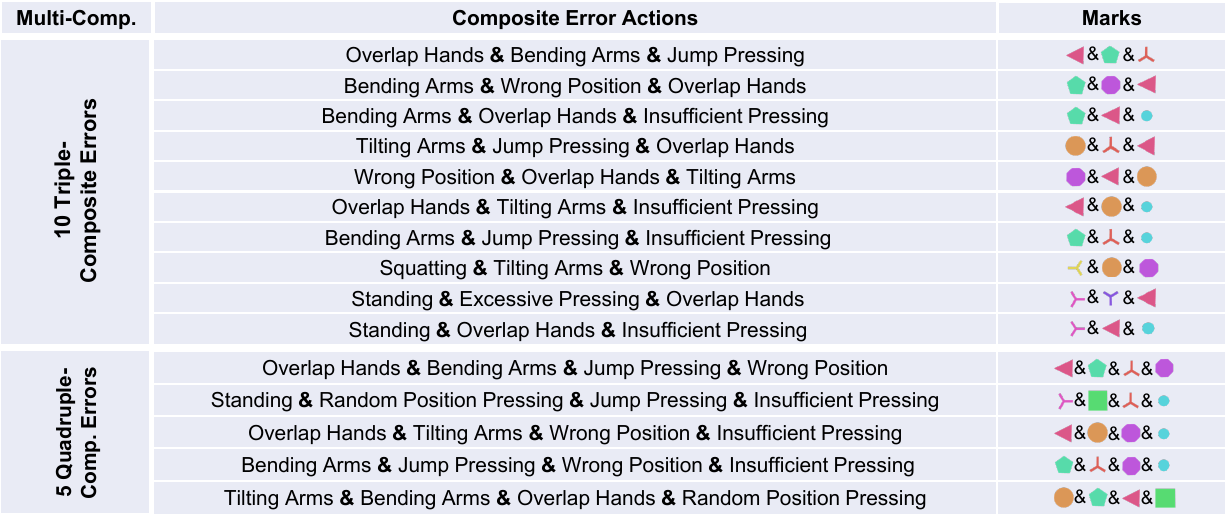}
	\caption{All combinations of the 10 triple- and 5 quadruple-composite error actions studied in this paper.}
	\label{ComActions15}
\end{figure}

\begin{figure} \centering
    \begin{center}
    	\includegraphics[width=.615\linewidth]{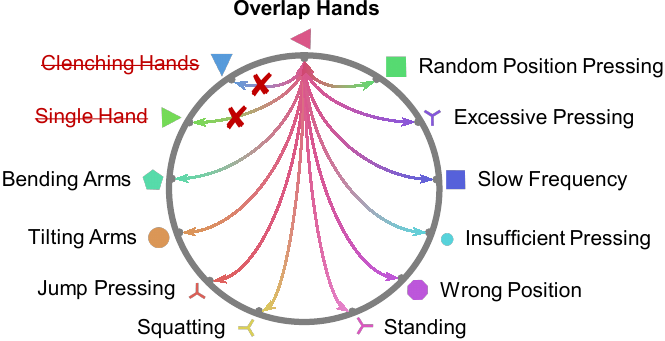}
    \end{center}
    \vspace{-8pt}
	\caption{The selection strategy of the composite error actions. In this case, \textit{Overlap Hands} is selected as the primary class, and two impossible co-occurrence combinations are deleted.}
	\label{Fig3}
\end{figure}

Considering the exponential growth of the total number of composite error actions (8191 classes for 13 single-error actions), this paper mainly focuses on paired combinations and several common multi-error combinations.
Based on the filtering strategy in Figure \ref{Fig3}, we remove 19 impossible combinations from 78 pairs ($C_{13}^{2} = 78$) and finally get 59 paired-composite error actions.
All deleted combinations have been confirmed by emergency doctors.
In addition, 10 triple errors and 5 quadruple errors are selected by these professional doctors based on actual experience.
All these 15 multi-composite error actions are listed in Figure \ref{ComActions15} in detail.
Finally, we built a label space containing 74 combination errors.

\noindent\textbf{Data Collection}. 
We build a video capture system with four high-resolution cameras to record the rescue process, as shown in Figure \ref{Fig1}(a).
In order to ensure the diversity of the dataset, we recruited 12 volunteers to participate in data collection.
Multiple participants enrich the visual feature diversity of the proposed dataset.
Three volunteers were assigned to \textit{Set-1}, and the other nine were assigned to \textit{Set-2}.
For the single-class actions in \textit{Set-1}, the number of performing times is 40. For composite error actions in \textit{Set-2}, the number of performing times is 8. 
Therefore, the number of videos in each error category is consistent, which provides a fair comparison in experiments.
All actions are carried out under the guidance of professional doctors to ensure the quality of each external cardiac compression action.

\begin{table}[t] \scriptsize \renewcommand \arraystretch{0.8}
\begin{center}
    \caption{Comparison with existing medical action analysis datasets.}
    \label{Tab1-Compare}
    \vspace{-5pt}
    \begin{tabular}{ccccccc}
        \toprule
        Dataset  & \#Actions & Modality & \#Videos & \#Views & Available \\
        \midrule
        FLS-ASU \cite{ZQ1_2011} & 1 & RGB & 28 & 2  & \textcolor[RGB]{192,0,0}{\ding{56}} \\
        Sharma \textit{et al.} \cite{Sharma} & 2 & RGB & 33 & 1 & \textcolor[RGB]{192,0,0}{\ding{56}} \\
        Bettadapura \textit{et al.} \cite{Bettadapura} & 3 & RGB & 64 & 2 & \textcolor[RGB]{192,0,0}{\ding{56}} \\
        Zia \textit{et al.} \cite{GIT-IJCARS-2016} & 2 & RGB & 104 & 1 & \textcolor[RGB]{192,0,0}{\ding{56}} \\
        Zhang \textit{et al.} \cite{ZQ2_2013} & 1 & RGB & 546 & 1 & \textcolor[RGB]{192,0,0}{\ding{56}} \\
        Chen \textit{et al.} \cite{ICME2015} & 3 & RGB & 720 & 2 & \textcolor[RGB]{192,0,0}{\ding{56}} \\
        MISTIC-SL \cite{JHU-MICCAI-2016} & 4 & RGB+Kinematics & 49 & 1 & \textcolor[RGB]{192,0,0}{\ding{56}} \\
        JIGSAWS \cite{JHU-JIGSAWS-2014} & 3 & RGB+Kinematics & 103 & 1 & \textcolor[RGB]{83,129,53}{\ding{52}} \\
        \midrule
        CPR-Coach (Ours) & 14+74 & RGB+Flow+Pose & 4,544 & 4 & \textcolor[RGB]{83,129,53}{\ding{52}} \\
        \bottomrule
    \end{tabular}
\end{center}
\end{table}

\begin{table} \footnotesize \tabcolsep=10pt \renewcommand \arraystretch{0.7}
    \caption{Summary of statistics of the CPR-Coach dataset.}
    \label{Tab2-Statistics}
    \vspace{-15pt}
\begin{center}
    \begin{tabular}{ccc}
        \toprule
        Item & Data\\
        \midrule
        Perspectives & 4  \\
        FPS & 25 \\
        Video Resolution & 4096×2160 (4K) \\ 
        \midrule
        Number of Participants & 12 \\
        Classes of Single-class Actions & 1+13=14 \\
        Classes of Composite Error Actions & 59+10+5=74 \\
        Frames (RGB) & 2,217,756 \\
        Frames (RGB+Flow) & 6,644,596 \\
        \midrule
        Videos & 4,544 \\
        Avg. Len. of Videos & 19.52s \\
        Storage Size & 449GB \\
        \bottomrule
    \end{tabular}
\end{center}
\end{table}

\begin{figure}
    \begin{center}
    	\includegraphics[width=.8\linewidth]{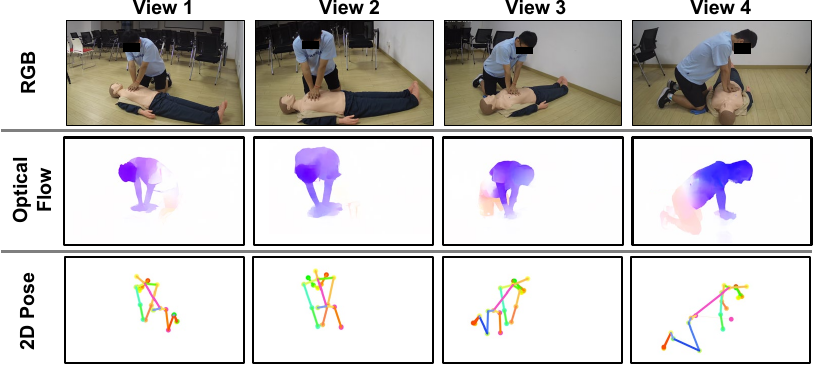}
    \end{center}
    \vspace{-10pt}
	\caption{The CPR-Coach dataset contains three types of modality information on four views.}
	\label{Fig5}
\end{figure}

\noindent\textbf{Dataset Statistics}.
Table \ref{Tab1-Compare} compares the proposed CPR-Coach dataset with existing medical action analysis
datasets.
The CPR-Coach dataset has surpassed existing datasets in terms of data scale, action granularity, and modal complexity.
Table \ref{Tab2-Statistics} summarizes the statistics of the CPR-Coach dataset.
It contains around 4.5K videos and 2.2M frames in total.
The storage size of the entire dataset is 449GB.
The CPR-Coach also provides optical flow images generated by the TV-L1 algorithm \cite{TV-L1} and 2D skeletons of the rescuer obtained by Alphapose \cite{Alphapose}.
Figure \ref{Fig5} shows three modality information from four perspectives: RGB frames, optical flow, and 2D poses.

\noindent\textbf{Supported Tasks}.
As the first multi-perspective dataset to explore fine-grained composite actions in medical scenarios, the CPR-Coach can support multiple studies.
Firstly, we can evaluate existing HAR models on fine-grained error recognition tasks on \textit{Set-1}.
Secondly, by taking \textit{Set-1} as the training set and \textit{Set-2} as the testing set, we can explore the composite error action recognition task under constrained supervision.
Thirdly, the influence of combining different perspectives and modes on the algorithm can be explored.
The following experiments follow these ideas.


\noindent\textbf{Ethics Issues}.
Studies in this paper only involve pure medical skill training and assessment. All video data were collected by volunteers with knowledge and consent.
Each participant signed a GDPR informed consent which allows the dataset to be publicly available for research purposes.

\section{ImagineNet}

\begin{figure}
    \begin{center}
    	\includegraphics[width=0.715\linewidth]{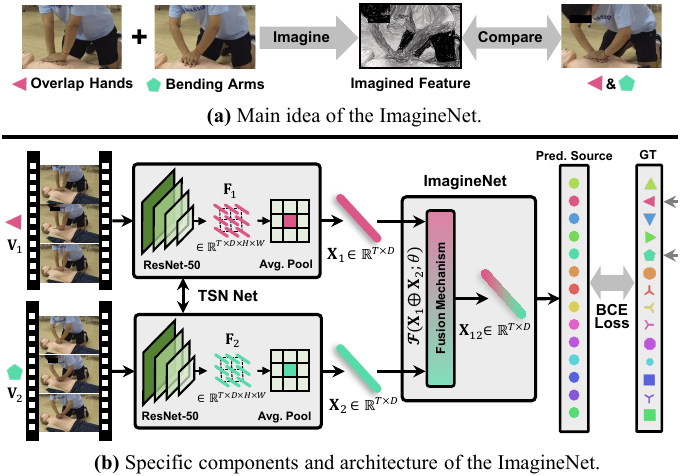}
    \end{center}
    \vspace{-8pt}
	\caption{(a) and (b) demonstrate the main idea and specific network architecture of the proposed ImagineNet, respectively. Two error actions \textit{Overleap Hands} and \textit{Bending Arms} are selected for visualization. 
    The ImagineNet simulates the thinking and judgment process of a real experienced coach concisely. 
    The knowledge base only includes single-class actions, while real applications will encounter unseen composite errors.
    }
	\label{Fig4_1}
\end{figure}

Humans have inherent learning and reasoning strengths.
Although a coach has not seen all possible combinations (8191 classes for 13 single-error actions in our settings), they can accurately determine the composite errors of the rescuer based on simple single-error cases.
Inspired from this, we proposed the ImagineNet, which can effectively handle such issues in Figure \ref{Fig1}(b).
Figure \ref{Fig4_1}(a) shows the main idea of the proposed human-cognition-inspired framework ImagineNet.
With restricted supervision training data, the \textit{Imagine} process can freely combine features to improve the multi-label recognition performance.
Taking the classic Temporal Segment Network (TSN) \cite{TSN2016ECCV} as the basic network, the detailed architecture of ImagineNet is shown in Figure \ref{Fig4_1}(b). 

\begin{figure}
    \begin{center}
    	\includegraphics[width=0.7\linewidth]{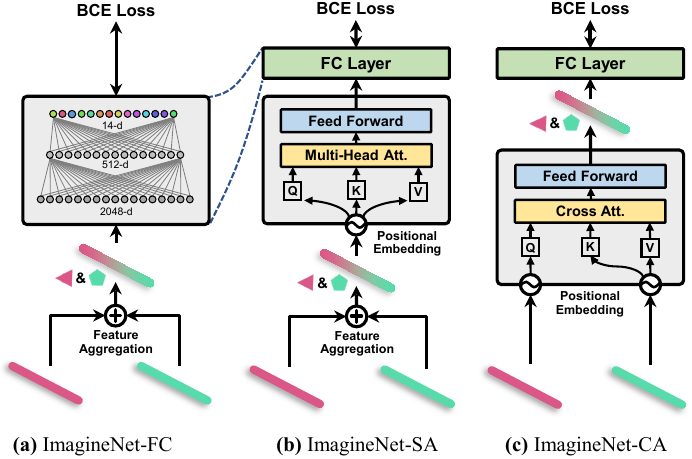}
    \end{center}
    \vspace{-8pt}
	\caption{Three feature fusion schemes proposed in this paper.
    Note that only two inputs are displayed for clarity.
    }
	\label{Fig4_2}
\end{figure}

The ImagineNet is divided into three stages: feature extraction, feature fusion, and loss computing.
Firstly, two video samples $(\mathbf{V}_1, C_1)$ and $(\mathbf{V}_2, C_2)$ are selected from \textit{Set-1} in the feature extraction phase.
Note that two videos $\mathbf{V}_1 = \{ I_i \}_{i=1}^{N_1}$ and $\mathbf{V}_2 = \{ I_i \}_{i=1}^{N_2}$ come from different classes, \textit{i.e.}, $C_1 \neq C_2, C \in \{1, \cdots, 13\}$.
$N_1$ and $N_2$ represent the total frames of two videos, respectively. $I_i$ denotes the $i$-th frame in the video.
The TSN model selects $T$ clips from raw videos for feature extraction.
After spatial average pooling, video features $\mathbf{X}_1 \in \mathbbm{R}^{T\times D}$ and $\mathbf{X}_2 \in \mathbbm{R}^{T\times D}$ are obtained, where $D$ denotes the dimension of the feature.
Secondly, in the feature fusion stage, two different features will be subsequently fused and generate $\mathbf{X}_{12} \in \mathbbm{R}^{T\times D}$.
This process is also expressed as $\mathbf{X}_{1} \oplus \mathbf{X}_{2}$.
We regard this feature fusion process as the \textit{Imagine} process. 
As illustrated in Figure \ref{Fig4_2}(a\&b\&c), this paper provides three feature fusion schemes to realize the imagination process: Fully-Connected Layer based fusion (FC), Self-Attention based fusion (SA), and Cross-Attention based fusion (CA).
Finally, in the loss computing stage, the Binary Cross Entropy (BCE) loss is adopted to measure the divergence between the predicted score and the Ground-Truth (GT) labels. Note that the GT labels are in the form of multi-hot encoding.

\subsection{Fusion Mechanisms of the ImagineNet}
Subfigures in Figure \ref{Fig4_2}(a\&b\&c) demonstrate three different feature fusion mechanisms: ImagineNet-FC, ImagineNet-SA, and ImagineNet-CA, respectively.
The formula representation is omitted in these figures for clarity. Two thick lines with different colors are adopted to represent two video features.

\noindent \textbf{ImagineNet-FC}. 
As shown in Figure \ref{Fig4_2}(a), the video features $\mathbf{X}_1$ and $\mathbf{X}_2$ are fused through the feature addition mechanism. Then a two-layer fully connected neural network maps the fusion feature $\mathbf{X}_{12}$ into predicted scores of 14 classes. 
This process is formulated as
\begin{equation}
    S_{FC} = \mathcal{F}_{FC} ( \mathbf{X}_{1} \oplus \mathbf{X}_{2} ; \bm{\theta}_{FC}),
    \label{Equ1}
\end{equation}
where $\mathcal{F}_{FC} (\cdot)$ denotes the neural network, and the plus sign $\oplus$ represents the feature aggregation strategy, which will be described in detail later. $\bm{\theta}_{FC}$ represents the trainable parameters of $\mathcal{F}_{FC} (\cdot)$.

The BCE loss function is selected for the network optimization: 
\begin{equation}
    \bm{\theta}^{*}_{FC} = \mathop{\arg\min}\limits_{\bm{\theta}_{FC}} BCE( S_{FC}, GT ),
\end{equation}
where $GT = onehot(C_1) \cup onehot(C_2)$ denotes the composite label in multi-hot encoding form.
All parameters are omitted in the subsequent statements for clarity.

\noindent \textbf{ImagineNet-SA}. 
The ImagineNet-SA adds a self-attention module based on the ImagineNet-FC, as shown in Figure \ref{Fig4_2}(b).
The motivation is to equip the ImagineNet with a stronger feature extraction and fusion capability to improve the generalization and reasoning ability.
The process is expressed as
\begin{equation}
    S_{SA} = \mathcal{F}_{FC} ( \mathcal{F}_{SA} ( \mathbf{X}_{1} \oplus \mathbf{X}_{2}) ),
    \label{Equ3}
\end{equation}
where $ \mathcal{F}_{SA} ( \cdot ) $ includes the self-attention and feed forward stages, $ \mathcal{F}_{FC} ( \cdot ) $ is the same as Equ.\ref{Equ1}.
By substituting $ \mathbf{X}_{12} $ for $ \mathbf{X}_{1} \oplus \mathbf{X}_{2} $, the self-attention mechanism is expressed as
\begin{equation}
    \mathbf{X}_{SA}^{'} = LN \left [ \mathbf{X}_{12} + softmax \left ( \frac{\mathbf{X}_{12} \mathbf{X}_{12} ^ {T}}{\sqrt{D}} \right ) \mathbf{X}_{12} \right ],
\end{equation}
and the feed forward layer 
\begin{equation}
    \mathbf{X}_{SA} = LN [ \mathbf{X}_{SA}^{'} + \mathcal{F}_{FFN} (\mathbf{X}_{SA}^{'}) ].
\end{equation}
Note that $D$ represents the dimension of video features and
\noindent $D=2048$ in TSN \cite{TSN2016ECCV}.
$LN[\cdot]$ denotes the LayerNorm operation. For clarity, the LayerNorm operation and residual links are omitted in Figure \ref{Fig4_2}(b\&c).

\noindent \textbf{ImagineNet-CA}. The structure of ImagineNet-CA is shown in Figure \ref{Fig4}(e). The main difference between ImagineNet-SA and ImagineNet-CA lies in the feature fusion strategy. 
Consistent with the above, the computing process is expressed as
\begin{equation}
    S_{CA} = \mathcal{F}_{FC} ( \mathcal{F}_{CA} ( \mathbf{X}_{1} , \mathbf{X}_{2}) ),
\end{equation}
where $\mathcal{F}_{CA} ( \cdot , \cdot )$ includes a cross-attention module and a feed forward layer.
The cross-attention mechanism integrates two video features from different classes:
\begin{equation}
    \mathbf{X}_{CA}^{'} = LN \left [ \mathbf{X}_{1} + softmax \left ( \frac{\mathbf{X}_{1} \mathbf{X}_{2} ^ {T}}{\sqrt{D}} \right ) \mathbf{X}_{2} \right ],
\end{equation}
and the feed forward layer
\begin{equation}
    \mathbf{X}_{CA} = LN [ \mathbf{X}_{CA}^{'} + \mathcal{F}_{FFN} (\mathbf{X}_{CA}^{'}) ].
\end{equation}

After defining three fusion mechanisms, we can instantiate three ImagineNets and compare their performance.
Three feature fusion mechanisms mentioned above are frameworks for implementing ImagineNet, while feature aggregation is a local operation denoted as $\oplus$. 
Effective feature aggregation methods can make full use of limited samples in \textit{Set-1}, thus improving the generalization performance under the setting of \textit{Single-class Training} \& \textit{Multi-class Testing} .

\subsection{Feature Aggregation Strategy}
\begin{figure} \centering
    \includegraphics[width=0.7\linewidth]{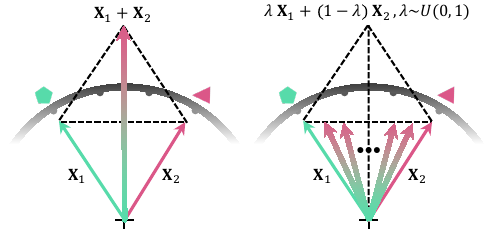}
    \caption{Visualization of the vanilla additive mechanism and the proposed weighted feature summation mechanism.}
    \label{Fig-Sum}
\end{figure}
The simplest way to instantiate $\oplus$ in ImagineNet-FC and -SA models is taking the summation of two features.
To increase the diversity of the aggregation process, we propose a random weighted summation mechanism based on the vanilla version.
As shown in Figure \ref{Fig-Sum}, the aggregated feature is expressed as
\begin{equation}
    \mathbf{X}_{12} = \lambda \mathbf{X}_{1} + (1-\lambda) \mathbf{X}_{2}, \lambda \sim U(0, 1),
\end{equation}
where $\lambda$ is a weight sampled from a uniform distribution $U(0, 1)$. 
This mechanism is able to enrich feature combinations. Consequently, the ImagineNet can \textit{Imagine} various combined situations given specific error actions.
The effectiveness of this concise technique is verified in ablation studies.
As representatives of feature aggregation methods, CBP \cite{CBP} and BLOCK \cite{Block} are selected for comparison.
Weighted summation, CBP, and BLOCK are denoted as Agg-1, Agg-2, and Agg-3, respectively.

\begin{sidewaystable} \scriptsize \tabcolsep=6.5pt \renewcommand \arraystretch{0.85}
    \caption{Single-class recognition performance of existing HAR models on CPR-Coach \textit{Set-1}. The first and second accuracy in each column are highlighted in \textbf{bold} and \underline{underlined}, respectively.}
    \vspace{-15pt}
    \label{Tab3-Single_Perfromance}
\begin{center}
    \begin{tabular}{cccccc|cc|cc|cc}
    \toprule
        \multirow{2.9}*{Model} & \multirow{2.9}*{Backbone} & \multirow{2.9}*{Config} & \multirow{2.9}*{Epoch} & \multirow{2.9}*{Modality} & \multirow{2.9}*{Pre-training} & \multicolumn{2}{c|}{CE Loss} & \multicolumn{2}{c|}{BCE Loss}  & \multicolumn{2}{c}{Multi-Margin Loss} \\
        \cmidrule(lr){7-12}
        ~ & ~ & ~ & ~ & ~ & ~ & Top-1 & Top-3  & Top-1 & Top-3 & Top-1 & Top-3 \\
        \midrule
        \multirow{3}*{TSN \cite{TSN2016ECCV}} & ResNet-50 & 1x1x8 & 50 & RGB & \ding{55}          & 0.8879 & 0.9940 & 0.8829 & \underline{0.9960} & 0.8502 & 0.9901 \\
        ~ & ResNet-50 & 1x1x8 & 50 & RGB & Kinetics-400                                   & 0.9067 & 0.9921 & 0.8919 & 0.9940 & 0.8690 & 0.9901 \\
        ~ & ResNet-50 & 1x1x8 & 50 & Flow & \ding{55}                                      & 0.7907 & 0.9603 & 0.8304 & 0.9851 & 0.7073 & 0.9355 \\
        \midrule
        TSM \cite{TSM2019ICCV} & ResNet-50 & 1x1x8 & 50 & RGB & \ding{55}   & 0.9067 & 0.9901 & 0.9325 & 0.9950 & 0.8433 & 0.9881 \\ 
        TRN \cite{TRN2018ECCV} & ResNet-50 & 1x1x8 & 50 & RGB & \ding{55}                  & 0.7827 & 0.9633 & 0.7421 & 0.9435 & 0.7431 & 0.9663 \\
        I3D \cite{i3d2017CVPR} & ResNet-50 & 32x2x1 & 50 & RGB &\ding{55}                  & 0.9692 & 0.9960 & 0.9117 & 0.9940 & 0.8591 & 0.9861 \\
        TPN \cite{TPN2020CVPR} & ResNet-50 & 8x8x1 & 50 & RGB & \ding{55}                  & \textbf{0.9802} & 0.9960 & 0.9087 & \textbf{0.9980} & 0.8720 & 0.9901 \\
        C3D \cite{C3D2015ICCV} & C3D & 16x1x1 & 50 & RGB & Sports1M                      & 0.9722 & 0.9931 & \textbf{0.9702} & 0.9931 & 0.8621 & 0.9802 \\
        TIN \cite{TIN2020AAAI} & ResNet-50 & 1x1x8 & 50 & RGB & \ding{55}                  & 0.8800 & 0.9901 & 0.7192 & 0.9335 & 0.8393 & 0.9861 \\
        SlowFast \cite{slowfast2019ICCV} & ResNet-50 & 4x16x1 & 256 & RGB & \ding{55}       & 0.8695 & 0.9734 & 0.8719 & 0.9781 & 0.8625 & 0.9688 \\
        TimeSFormer \cite{timesformer2021ICML} & ViT & 8x32x1 & 50 & RGB & \ding{55}     & 0.8879 & 0.9921 & 0.8998 & 0.9940 & 0.8462 & 0.9762 \\
        \midrule
        ST-GCN \cite{ST-GCN} & ST-GCN & 1x1x300 & 50 & Pose & \ding{55} & 0.9246 & \textbf{0.9970} & 0.9187 & 0.9881 & 0.9196 & \textbf{0.9970} \\
        PoseC3D \cite{PoseC3D} & ResNet3D-50 & 1x1x300 & 240 & Pose & \ding{55} & 0.9208 & 0.9922 & 0.9035 & 0.9715 & 0.8837 & 0.9606 \\
        \midrule
        \multirow{2}*{Two-Stream \cite{Two-Stream}} & TSN+TSN\_Flow & Late-Fusion & 50 &  RGB{\scriptsize +}Flow & \ding{55}                               & 0.9533 & 0.9891 & 0.9479 & 0.9825 & \underline{0.9296} & 0.9802 \\
        ~ & TSN{\scriptsize +}ST-GCN & Late-Fusion & 50 & RGB+Pose & \ding{55} & \underline{0.9782} & \underline{0.9962} & \underline{0.9608} & 0.9941 & \textbf{0.9692} & \underline{0.9960} \\
        \bottomrule
    \end{tabular}
\end{center}
    \caption{Composite error action recognition performance on \textit{Set-2} by direct migration. Only the results of four models in RGB and pose modality are reported due to the limited space. Significant performance degradation can be observed compared to the results in Table \ref{Tab3-Single_Perfromance}.}
    \vspace{-15pt}
\begin{center}
    \begin{tabular}{cccc|cc|cc|cc}
        \toprule
        \multirow{2.9}*{Model} & \multirow{2.9}*{Config} & \multirow{2.9}*{Modality} & \multirow{2.9}*{Pre-training} & \multicolumn{2}{c|}{CE Loss}  & \multicolumn{2}{c|}{BCE Loss} & \multicolumn{2}{c}{Multi-Margin Loss} \\
        \cmidrule(lr){5-10}
        ~ & ~ & ~ & ~ & mAP & mmit mAP & mAP & mmit mAP & mAP & mmit mAP \\
        \midrule
        TSN \cite{TSN2016ECCV} & 1x1x8 & RGB & Kinetics-400 & 0.5598 & 0.6143 & 0.4627 & 0.5629 & 0.4838 & 0.5579 \\
        TPN \cite{TPN2020CVPR} & 8x8x1 & RGB & \ding{55} & \textbf{0.6250} & \textbf{0.7016} & 0.5201 & 0.6102 & 0.5457 & 0.6247 \\
        TSM \cite{TSM2019ICCV} & 1x1x8 & RGB & \ding{55} & 0.5662 & 0.6618 & \underline{0.5721} & \underline{0.6688} & \underline{0.5470} & \underline{0.6255} \\
        ST-GCN \cite{ST-GCN} & 1x1x300 & Pose & \ding{55} & \underline{0.5776} & \underline{0.6692} & \textbf{0.5868} & \textbf{0.6865} & \textbf{0.5874} & \textbf{0.6719} \\
        \bottomrule
    \end{tabular}
\end{center}
    \label{Tab4-Composite_Performance}
\end{sidewaystable}

\subsection{Inference of the ImagineNet}
Figure \ref{Fig4_1}(b) only demonstrates the training process of the ImagineNet. 
It can be found that ImagineNet requires two video features $\mathbf{X}_{1}$ and $\mathbf{X}_{2}$ as inputs during training.
However, there is only one input video feature of the composite error action during inference.
To resolve this mismatch issue, this paper directly adopts the replication method to fill the input.
Although the cross-attention in ImagineNet-CA degenerates into the self-attention in ImagineNet-SA during inference, different training process leads to different recognition performance. 
The two models are still comparable, and the experimental results confirm this analysis.

\begin{figure}[t]
    \begin{center}
    	\includegraphics[width=0.8\linewidth]{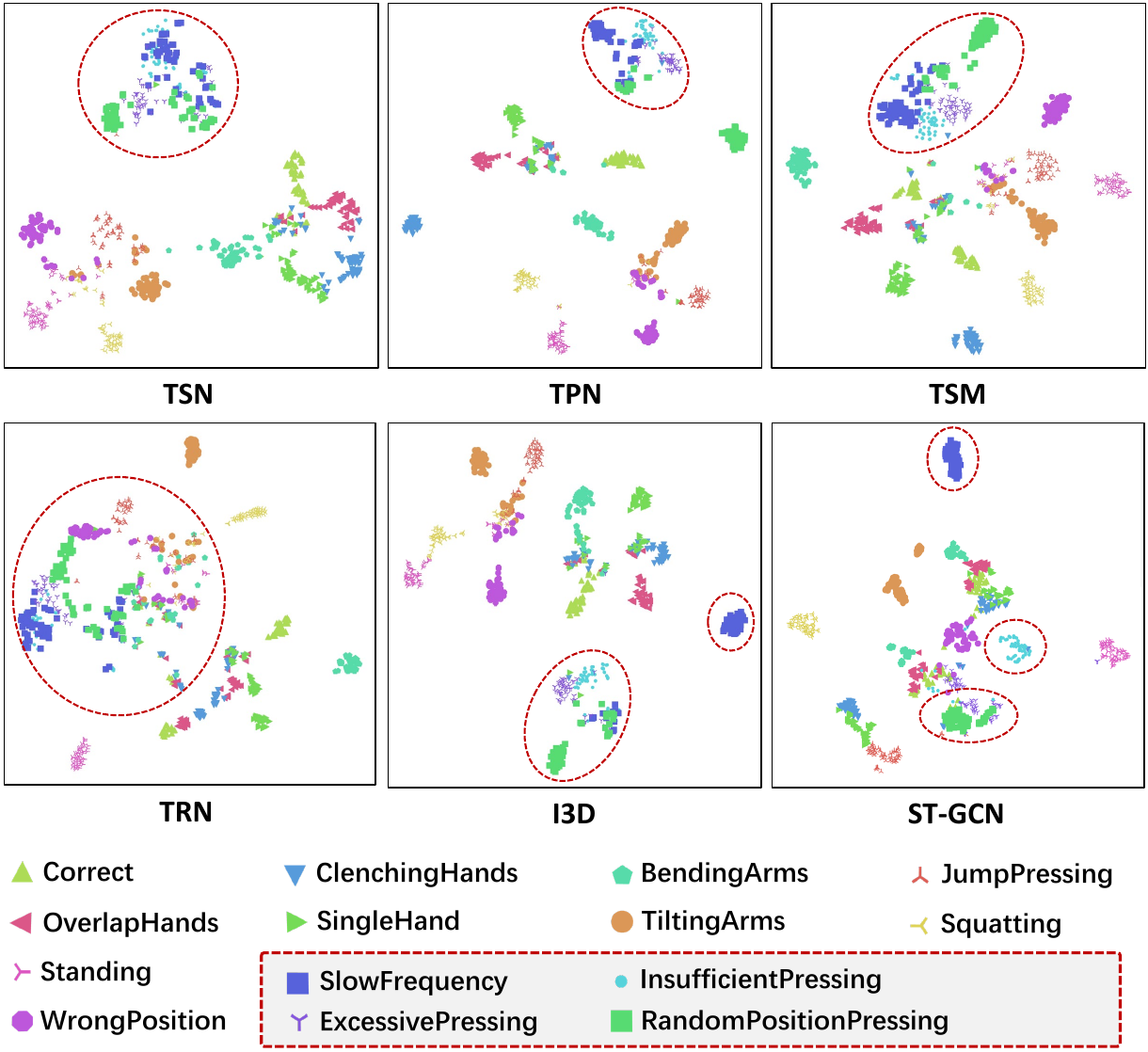}
    \end{center}
    \vspace{-8pt}
	\caption{Visualization of the action features through t-SNE. The red box in the legend highlights four confusing classes. We use red circles to highlight these four classes of scatters in figures to compare the performance of these networks more clearly.}
	\label{Fig6}
\end{figure}

\section{Experiments}

\subsection{Action Recognition on CPR-Coach Set-1}
Compared with traditional HAR datasets, the CPR-Coach focuses on distinguishing subtle errors in CPR.
In Figure \ref{Fig2}, it is difficult to find the nuances of these actions.
CPR-Coach puts forward higher requirements for the action recognition models.
Therefore, we take \textit{Set-1} of the CPR-Coach as a benchmark and conduct single-error recognition experiments on existing HAR models.
60\% of \textit{Set-1} is used for training and 40\% for testing.
Table \ref{Tab3-Single_Perfromance} summarizes the detailed settings and Top-1\&3 accuracy of the models. Figure \ref{Fig6} visualizes some features generated by these models through the t-SNE algorithm \cite{t-SNE}.

\noindent \textbf{Implementation Details}.
Default configurations in original papers of TSN \cite{TSN2016ECCV}, Two-Stream \cite{Two-Stream}, TSM \cite{TSM2019ICCV},TPN \cite{TPN2020CVPR}, TRN \cite{TRN2018ECCV}, I3D \cite{i3d2017CVPR}, C3D \cite{C3D2015ICCV}, TIN \cite{TIN2020AAAI}, SlowFast \cite{slowfast2019ICCV}, TimeSFormer \cite{timesformer2021ICML}, ST-GCN \cite{ST-GCN} and PoseC3D \cite{PoseC3D} are adopted in this study.
All models are trained for 50 epochs through the SGD optimizer, except the SlowFast \cite{slowfast2019ICCV} with the Cosine Annealing optimizer for 256 epochs and the PoseC3D \cite{PoseC3D} for 240 epochs.
The network input size is 224$\times$224, while the coordinates of 2D poses remain unchanged at 4096$\times$2160.
All models are built on Pytorch and implemented on a system with an Intel Xeon E5-2698 V4@2.20GHz CPU and an NVIDIA Tesla V100 GPU.
Cross Entropy (CE), BCE, and Multi-Margin losses are adopted to compare the performance comprehensively.

\noindent \textbf{Performance Analysis}. 
Results in Table \ref{Tab3-Single_Perfromance} suggest that these models can effectively handle single error classification tasks.
Different networks and different loss function combinations will affect the final classification performance. 
Better results can be achieved under the CE loss setting, which is mainly caused by the strong assumption that the CE loss function has mutual exclusion between classes.
In all tested models, the Two-Stream framework can achieve stable performance under different loss functions. 
Results on Two-Stream framework show that the fusion of different information modes contributes to performance improvement. 
In Figure \ref{Fig6}, the scatters of four confusing classes are very close in TSN, TPN, and TSM, while the I3D and ST-GCN that pay more attention to temporal information can handle these situations well.
In addition, results show that the latest PoseC3D \cite{PoseC3D} did not outperform the early ST-GCN model.
This may be mainly caused by two reasons: on the one hand, the complexity of PoseC3D is much higher than that of ST-GCN, resulting in overfitting on the single error recognition dataset; On the other hand, 
the PoseC3D  stacks 2D keypoints to form a 3D heatmap volume for action recognition.
However, because external cardiac compression is a repetitive and cyclical action, stacking 2D keypoints will destroy circulation.
Therefore, the performance of PoseC3D is inferior to the ST-GCN.
In subsequent experiments and analysis, ST-GCN is selected as the backbone network to ensure simplicity and reproducibility, rather than the PoseC3D.
Next, we will explore composite error performance on these models.

\subsection{Composite Error Action Recognition on Set-2}
Taking \textit{Set-1} as the training set and \textit{Set-2} as the testing set, we can simulate the real CPR assessment.
A naive approach is directly migrating the pre-trained model in single-class task to the composite error recognition task.
Table \ref{Tab4-Composite_Performance} summarizes the performance of four selected models. 
All three losses cannot handle the huge gap between the two tasks. The sharp decline in performance indicates that the new task has exceeded the representation capability of original models. 
It should be noted that the core contribution of this paper is not to create a novel HAR model but to build a better composite error detector through existing models.
Results in Table \ref{Tab3-Single_Perfromance} show that SOTA algorithms have not demonstrated impressive performance in the single-error recognition task.
Therefore, we adopt classic models such as TSN \cite{TSN2016ECCV}, TSM \cite{TSM2019ICCV}, and ST-GCN \cite{ST-GCN} to instantiate ImagineNets for ensuring the reproducibility and stability, instead of those sophisticated methods such as TimeSFormer and PoseC3D.
Next, the deployment details and results of the ImagineNet will be introduced.

\noindent \textbf{Implementation Details}.
All ImagineNet models are trained for 60 epochs through the SGD optimizer. The learning rate is set to 0.001 initially and attenuated by 0.1 at 20 and 40-th epochs.
The temporal length $T$ is set to 8.
Only the models trained with CE loss are explored.

\noindent \textbf{Evaluation Metrics}.
The \textit{mAP} and \textit{mmit mAP} metrics are adopted in this paper to evaluate the composite error action recognition performance. 
The \textit{mAP} refers to the \textit{macro mAP} in \cite{MMIT}, which denotes the average of the mean average precision for each class
\begin{equation}
    mAP = \frac{\sum_{i=1}^{C} AP_i }{C}.
    \label{Equ1}
\end{equation}
The \textit{mmit mAP} refers to the \textit{micro mAP} in \cite{MMIT}, which denotes the mean average precision over all videos
\begin{equation}
    mmit\ mAP = \frac{\sum_{j=1}^{N} AP_j }{N}.
    \label{Equ2}
\end{equation}
Note that $AP_i$ denotes the average precision over the $i$-th class, while $AP_j$ denotes the average precision for the $j$-th sample.
In the CPR-Coach dataset, numbers of samples in each classes are relatively balanced, so it can be found that the value of \textit{mmit mAP} is generally higher than that of \textit{mmit mAP}.

\begin{table} \footnotesize \renewcommand \arraystretch{0.7}
    \caption{Performance comparison between direct migration and ImagineNet-FC. All model settings are consistent with Table \ref{Tab4-Composite_Performance}. }
    \label{Tab5-Comparison}
    \vspace{-15pt}
    \begin{center}
    \begin{tabular}{c|cc|cc}
        \toprule
        Model & mAP & $\Delta$ & mmit mAP & $\Delta$ \\
        \midrule
        TSN \cite{TSN2016ECCV} & 0.5598 & --- & 0.6143 & --- \\
        \textit{w}/ ImagineNet-FC
         & \textbf{0.6259} & $\uparrow$ 6.61\% & \textbf{0.6893} & $\uparrow$ 8.50\% \\
        \midrule
        TPN \cite{TPN2020CVPR} & 0.6250 & --- & 0.7016 & --- \\
        \textit{w}/ ImagineNet-FC & \textbf{0.7094} & $\uparrow$ 8.44\% & \textbf{0.7620} & $\uparrow$  6.04\% \\
        \midrule
        TSM \cite{TSM2019ICCV} & 0.5662 & --- & 0.6618 & --- \\
        \textit{w}/ ImagineNet-FC & \textbf{0.7053} & $\uparrow$ 13.91\% & \textbf{0.7566} & $\uparrow$ 9.48\% \\
        \midrule
        ST-GCN \cite{ST-GCN} & 0.5776 & --- & 0.6692 & --- \\
        \textit{w}/ ImagineNet-FC & \textbf{0.6404} & $\uparrow$ 6.28\% & \textbf{0.7115} & $\uparrow$ 4.23\% \\
        \bottomrule
    \end{tabular}
    \end{center}
\end{table}

\begin{table}[t] \footnotesize \renewcommand \arraystretch{0.7}
    \caption{Performance and FLOPs comparison of the proposed three ImagineNet models and their variants based on the TSN.}
    \label{Tab6-FLOPs}
    \vspace{-15pt}
\begin{center}
    \begin{tabular}{ccccc}
        \toprule
        Model & Variants & GFLOPs & mAP & mmit mAP \\
        \midrule
        ImagineNet-FC & FC & 0.001 & 0.6259 & 0.6893\\
        \midrule
        \multirow{4}*{ImagineNet-SA} & SA & 0.068 & 0.6426 & 0.7049 \\
        ~ & SAx2 & 0.136 & \textbf{0.6450} & \textbf{0.7131} \\
        ~ & SAx3 & 0.203 & \underline{0.6436} & \underline{0.7086} \\
        ~ & \textit{w/o} PosEmb & 0.068 & 0.6305 & 0.6906 \\
        \midrule
        \multirow{4}*{ImagineNet-CA} & CA & 0.068 & 0.6307 & 0.6933 \\
        ~ & CA+SA & 0.136 & \textbf{0.6347} & \underline{0.7005} \\
        ~ & CA+SAx2 & 0.203 & \underline{0.6335} & \textbf{0.7046} \\
        ~ & \textit{w/o} PosEmb & 0.068 & 0.6281 & 0.6953 \\
        \bottomrule
    \end{tabular}
\end{center}
\end{table}

\noindent \textbf{Quantitative Analysis}. 
Table \ref{Tab5-Comparison} compares the ImagineNet-FC model with the vanilla migration method. 
Through the \textit{Imagine} mechanism, the ImagineNet-FC significantly improves the composite error recognition performance under restricted supervision, regardless of the input modality.
In particular, the ImagineNet-FC brings 13.91\% \textit{mAP} and 9.48\% \textit{mmit mAP} improvement on TSM. 
The performance and computational complexity of ImagineNet-SA, -CA, and their variants based on the TSN model are summarized in Table \ref{Tab6-FLOPs}. Same settings are also adopted to the TSM model, and the results are listed in Table \ref{TSM-FLOPs}.
The results reveal that the ImagineNet-SA outperforms the other two models, while
the CA mechanism does not improve performance as well as SA.
More layers and computational complexity will lead to overfitting. 
The \textit{Positional Embedding} module is essential in ImagineNets because chronological information is indispensable for distinguishing these fine-grained error actions.
In Figure \ref{Fig8}, we explore the relationship between the number of error combinations and the final performance on \textit{Set-2}.
The \textit{mmit\_mAP} of ImagineNet-FC gradually decreases as the number of composite errors increases, which is consistent with our intuition that more complex error combinations imply higher task difficulty.
In previous experiments, we only adopted single-modal backbone networks. In order to explore the upper bound of the composite error recognition performance, we additionally adopte the MMNet \cite{MMNet} as the multi-mode backbone to extract video features. The performance gain is listed in Table \ref{Tab_MMNet_Performance}. 
Results reveal that the video backbone model with stronger representation ability is more able to benefit from the proposed ImagineNet, which also demonstrates the effectiveness of the ImagineNet.

\begin{table} \footnotesize \renewcommand \arraystretch{0.7}
    \caption{Performance and FLOPs comparison of the proposed three ImagineNet models and their variants based on the TSM.}
    \label{TSM-FLOPs}
    \vspace{-15pt}
\begin{center}
    \begin{tabular}{ccccc}
        \toprule
        Model & Variants & GFLOPs & mAP & mmit mAP \\
        \midrule
        ImagineNet-FC & FC & 0.001 & 0.7053 & 0.7566\\
        \midrule
        \multirow{4}*{ImagineNet-SA} & SA & 0.068 & \textbf{0.7011} & \underline{0.7630} \\
        ~ & SAx2 & 0.136 & \underline{0.7007} & \textbf{0.7656} \\
        ~ & SAx3 & 0.203 & 0.6995 & 0.7572 \\
        ~ & \textit{w/o} PosEmb & 0.068 & 0.6822 & 0.7593 \\
        \midrule
        \multirow{4}*{ImagineNet-CA} & CA & 0.068 & \underline{0.6752} & 0.7346 \\
        ~ & CA+SA & 0.136 & \textbf{0.6766} & \textbf{0.7406} \\
        ~ & CA+SAx2 & 0.203 & 0.6728 & \underline{0.7377} \\
        ~ & \textit{w/o} PosEmb & 0.068 & 0.6725 & 0.7339 \\
        \bottomrule
    \end{tabular}
\end{center}
\end{table}

\begin{figure}
    \begin{center}
        \subfloat[TSM]{ 
            \includegraphics[width=0.44\linewidth]{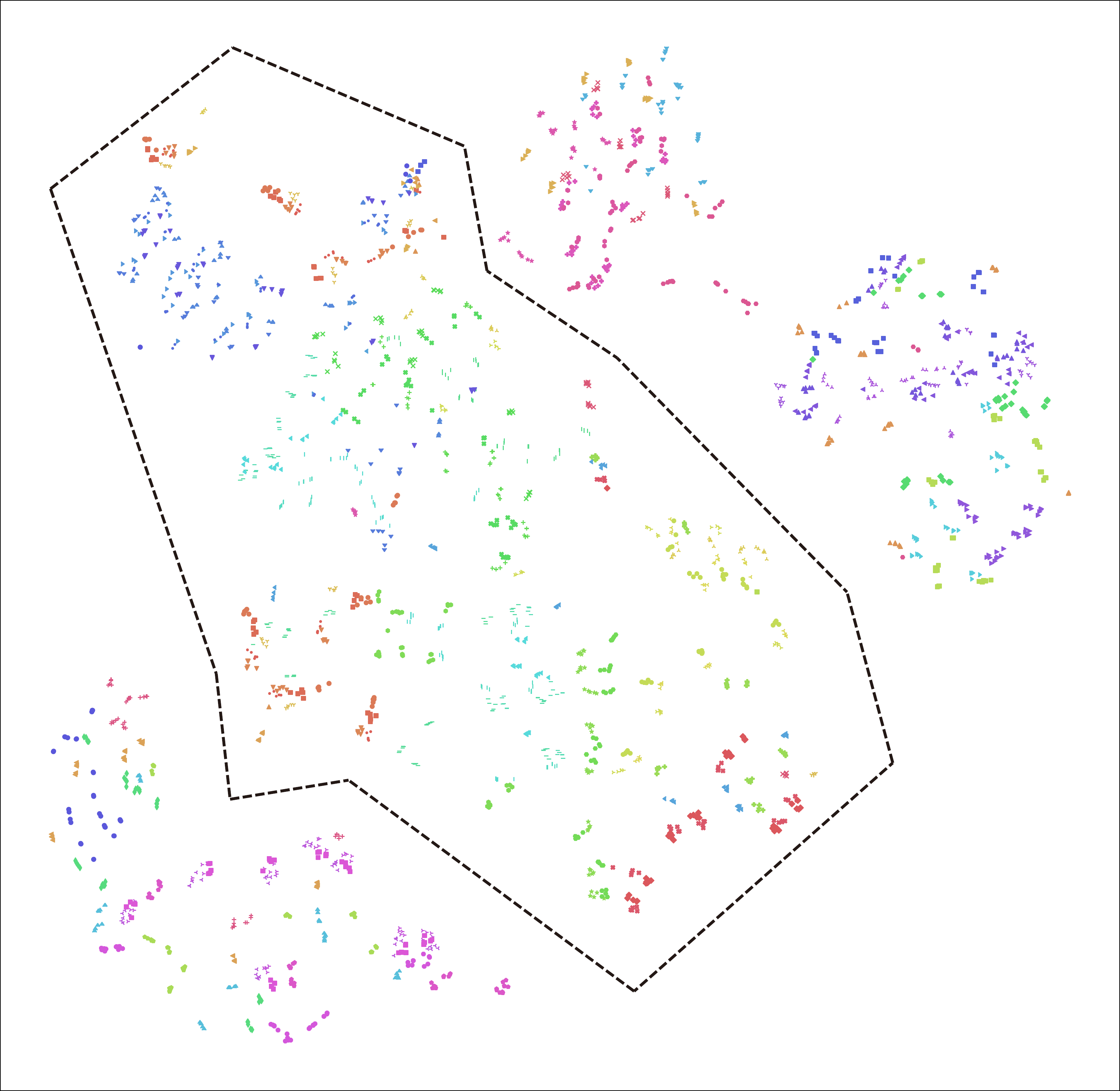}
        }
        \subfloat[TSM \textit{w}/ ImagineNet-FC]{ 
            \includegraphics[width=0.44\linewidth]{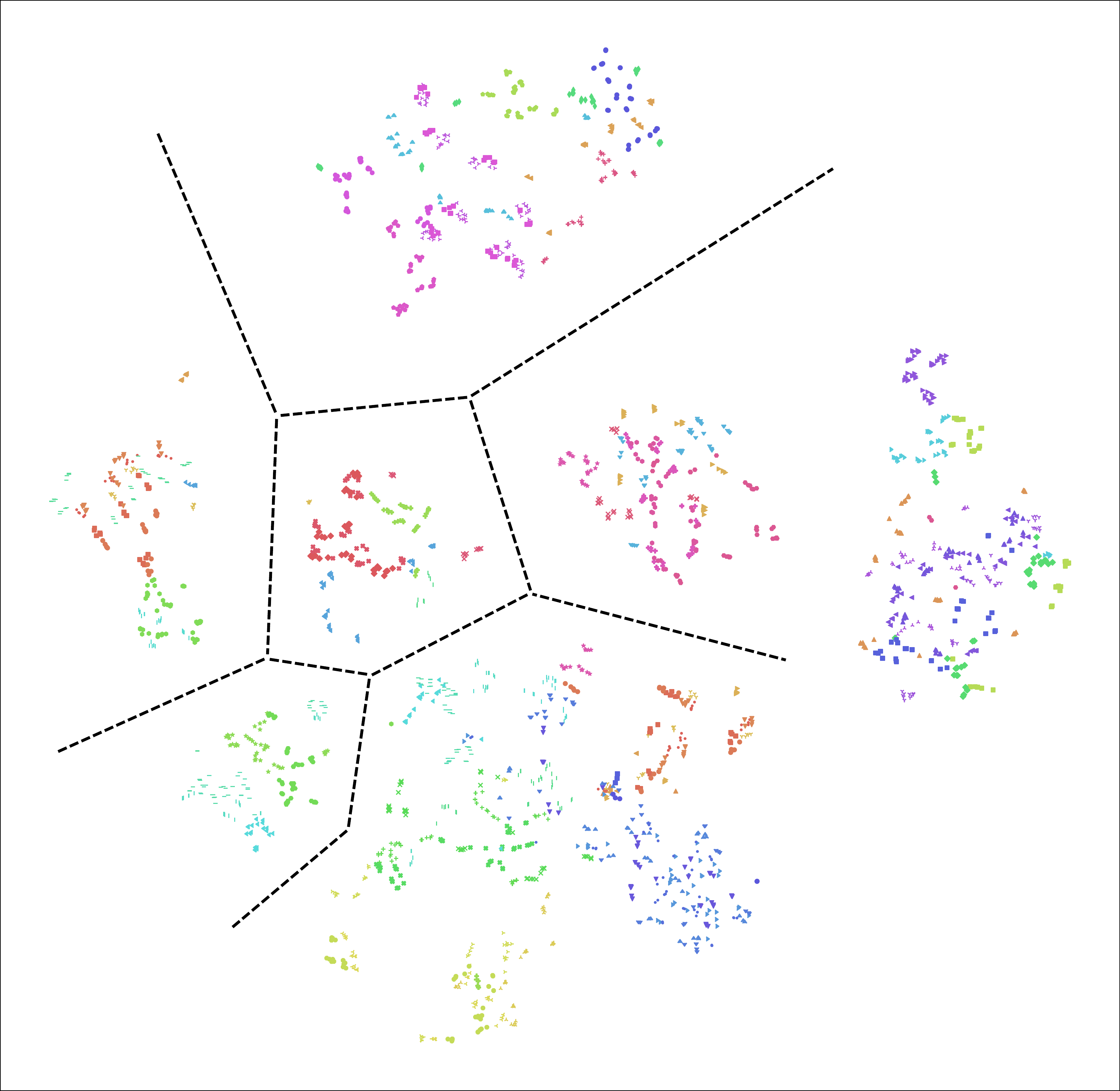}
        }
    \vspace{-10pt}
    \end{center}
    \begin{center}
        \subfloat[TSN]{ 
            \includegraphics[width=0.44\linewidth]{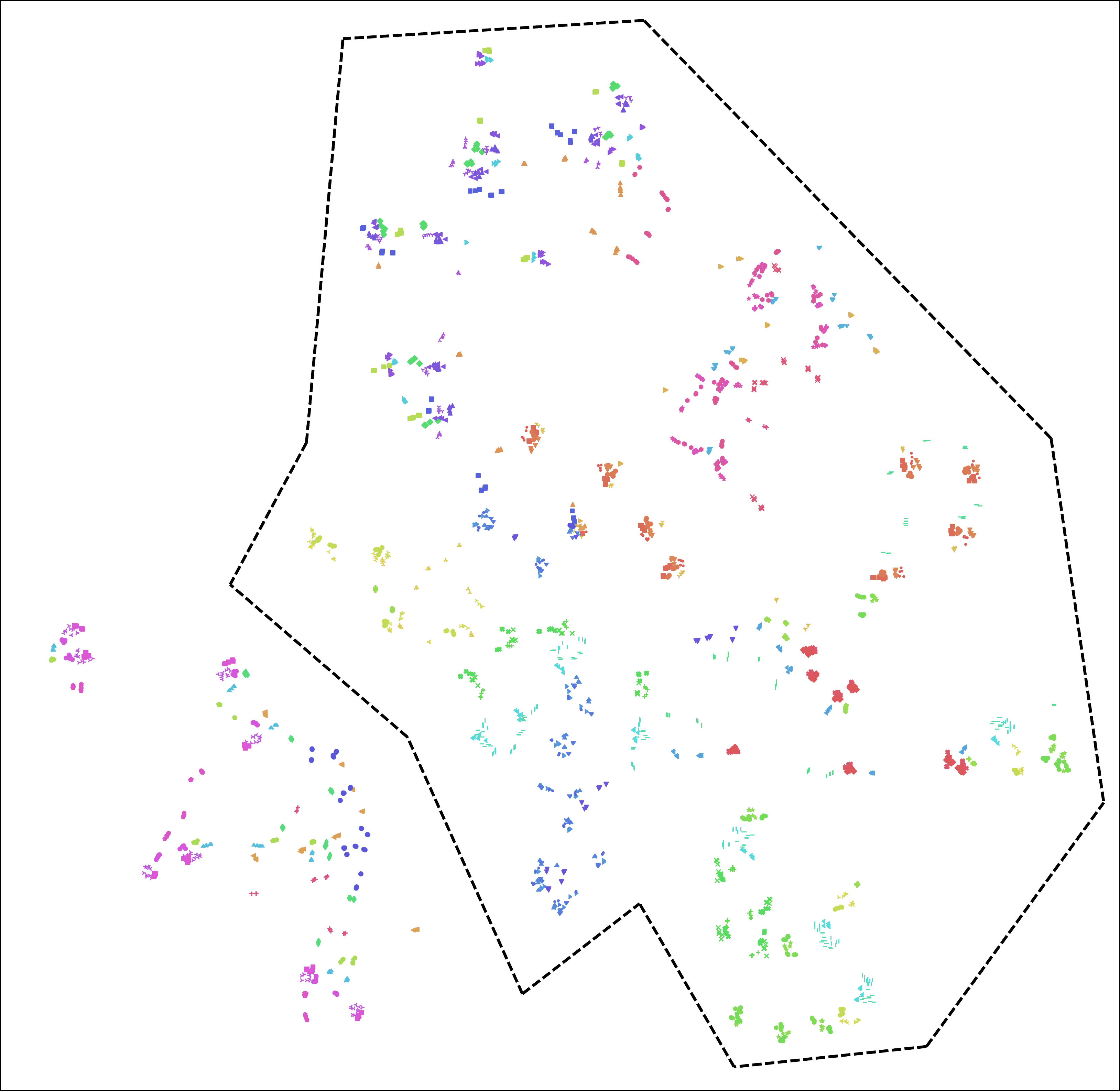}
        }
        \subfloat[TSN \textit{w}/ ImagineNet-FC]{ 
            \includegraphics[width=0.44\linewidth]{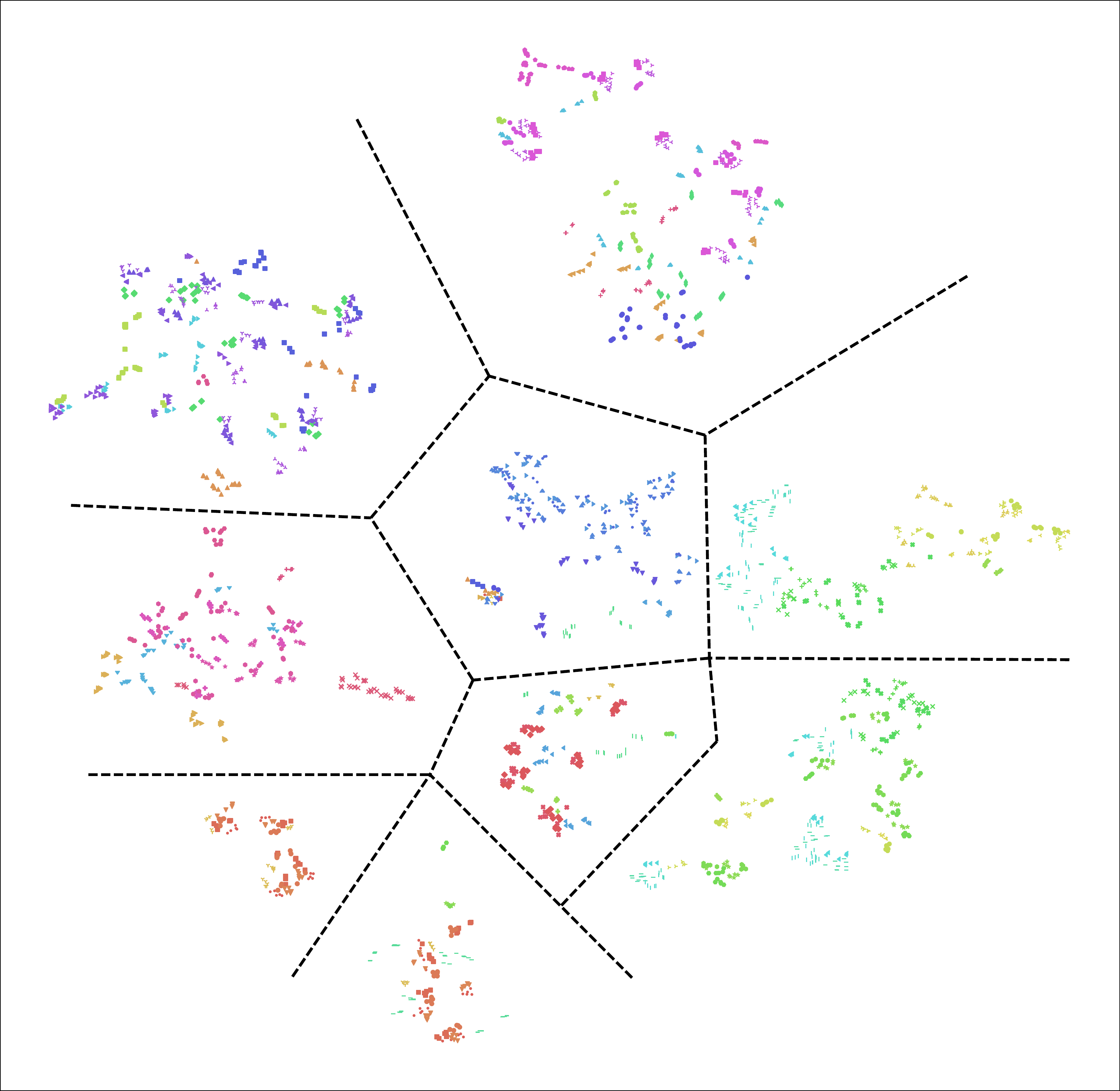}
        }
    \vspace{-10pt}
    \end{center}
    \begin{center}
        \subfloat[TPN]{ 
            \includegraphics[width=0.44\linewidth]{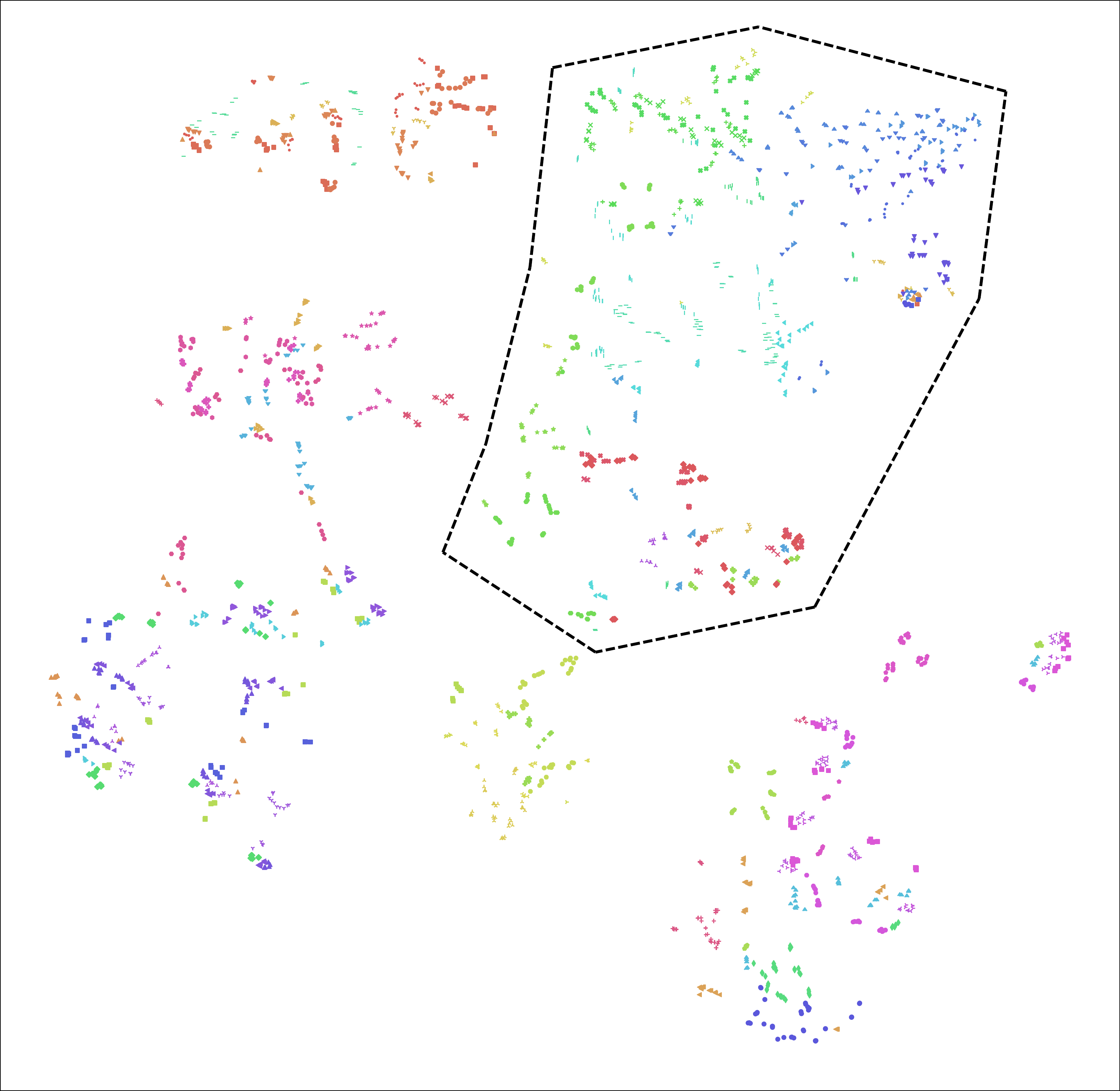}
        }
        \subfloat[TPN \textit{w}/ ImagineNet-FC]{ 
            \includegraphics[width=0.44\linewidth]{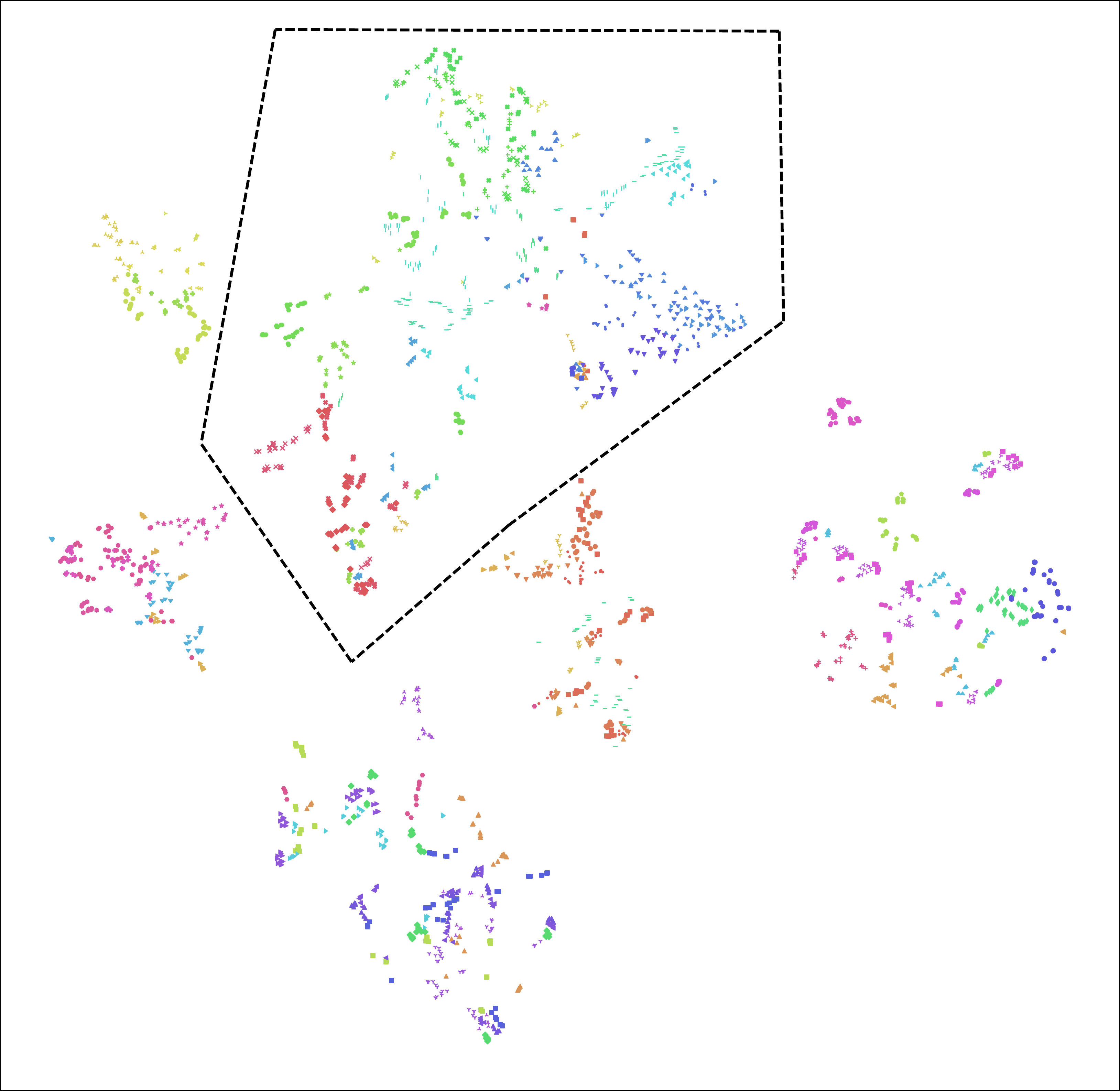}
        }
    \vspace{-10pt}
    \end{center}
    \caption{Feature visualization comparison via t-SNE on \textit{Set-2}. Black auxiliary lines are marked for clarity.}
    \label{Fig9}
\end{figure}

\begin{table}[t] \footnotesize \renewcommand \arraystretch{0.7}
    \caption{Performance testing based on MMNet, which has stronger representation ability in video feature extraction.}
    \label{Tab_MMNet_Performance}
    \vspace{-15pt}
    \begin{center}
        \begin{tabular}{cc|cc|cc}
        \toprule 
        Backbone & Variants & mAP & $\Delta$ & mmit mAP & $\Delta$ \\
        \midrule
        MMNet & Direct Migration & 0.6527 & --- & 0.7085 & --- \\
        \midrule
        \multirow{3}*{MMNet} & \textit{w}/ ImagineNet-FC & 0.7385 & $\uparrow$ 8.58\% & \underline{0.7716} & $\uparrow$ 6.31\% \\
        ~ & \textit{w}/ ImagineNet-SA & \textbf{0.7449} & $\uparrow$ 9.22\% & \textbf{0.7839} & $\uparrow$ 7.54\% \\
        ~ & \textit{w}/ ImagineNet-CA & \underline{0.7401} & $\uparrow$ 8.74\% & 0.7696 & $\uparrow$ 6.11\% \\
        \bottomrule
        \end{tabular}
    \end{center}
\end{table}
\noindent \textbf{Qualitative Analysis}. 
To explore how the proposed ImagineNet impacts the network, we visualize and compare the features generated by TSN, TSM, TPN and their ImagineNet-FC variant models on \textit{Set-2} in Figure \ref{Fig9}.
Macroscopically, features obtained by the direct migration method are messy,
while the ImagineNet can help the network reduce intra-class distance and expand inter-class distance.
These improvements on t-SNE feature maps correspond to the performance gains in Table \ref{Tab5-Comparison}.
The enhancement of feature clustering confirms the effectiveness of the proposed ImagineNet.

\subsection{Combination of Perspectives}
\begin{figure}[t]
    \begin{center}
        \includegraphics[width=.9\linewidth]{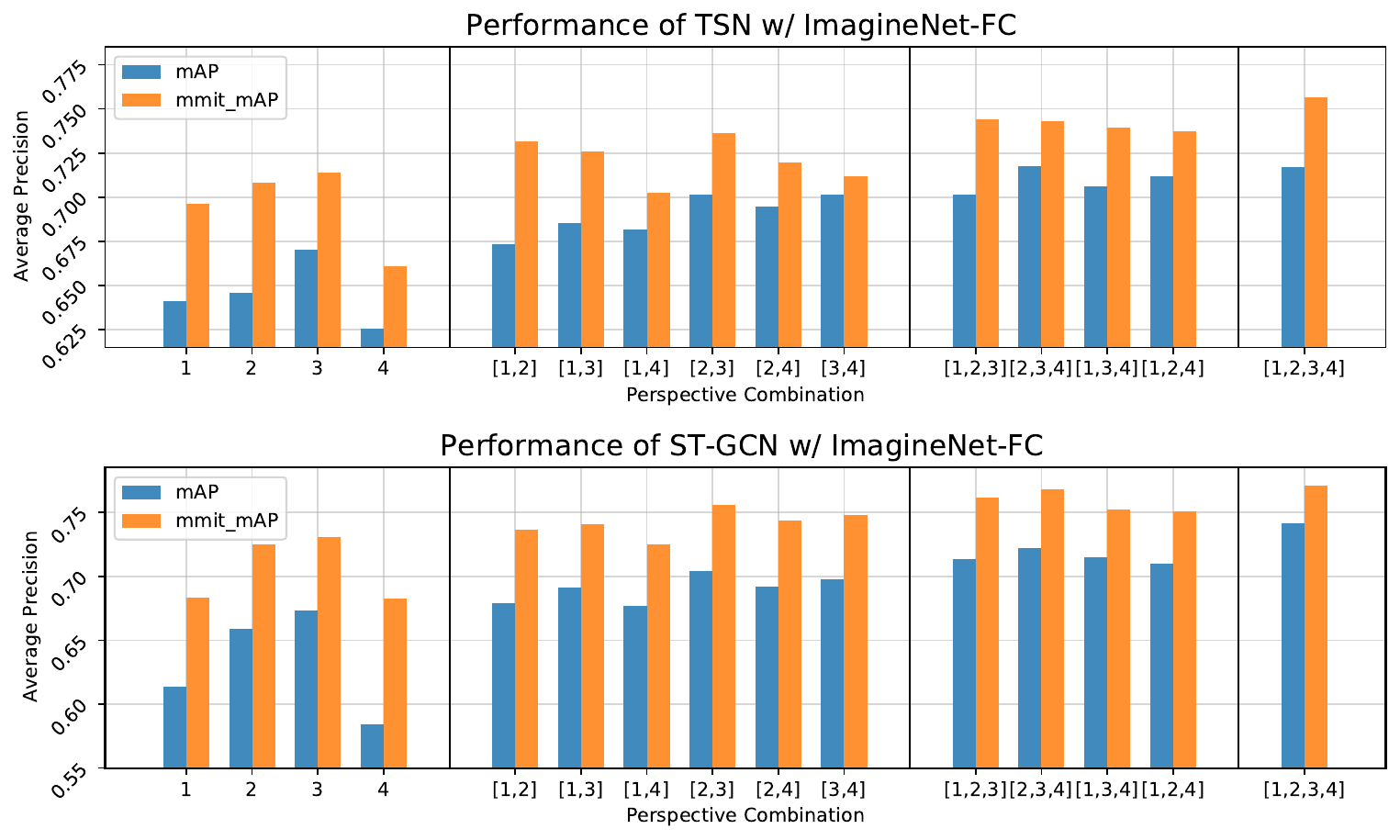}
    \end{center}
    \vspace{-10pt}
    \caption{Performance of combining different perspectives.
    Different numbers of views are grouped by black dividing lines.
    }
    \label{Fig11}
\end{figure}

\begin{figure}[t]
    \begin{center}
        \includegraphics[width=0.8\linewidth]{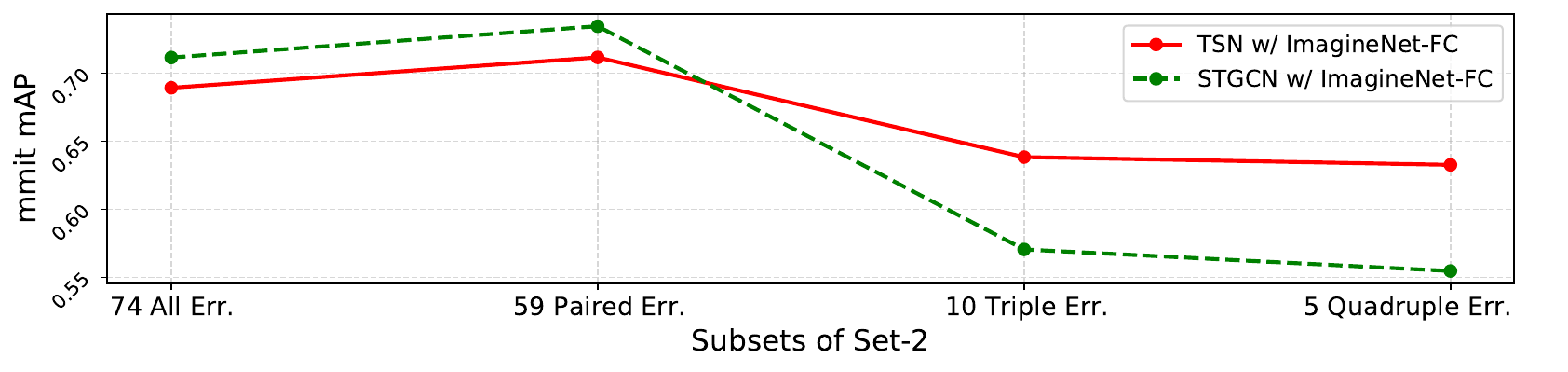}
    \end{center}
    \vspace{-10pt}
    \caption{\textit{mmit mAP} Performance on different subsets of \textit{Set-2}.
    }
    \label{Fig8}
\end{figure}

As shown in Figure \ref{Fig1}(a) and Figure \ref{Fig5}, the proposed video capture system includes four views. It is not practical to use all perspectives in actual deployment, which will cause too much redundant computation.
Four-perspective settings can help us discover the best combination and achieve the optimal performance-computation trade-off.
Consistent with the paradigm in Table \ref{Tab4-Composite_Performance}, we evaluate the performance of the ImagineNet-FC on all different perspectives combinations. 
Results are shown in Figure \ref{Fig11}. 
Overall, the performance increases with combing more perspectives. 
Perspective \#3 provides more valuable information, while \#4 is the opposite.
This discovery is of great value for subsequent system optimization and actual deployment.

\subsection{Ablation Studies}
Ablation studies are conducted to explore the effectiveness of feature aggregation strategies.
Table \ref{Tab7-Ablation} summarizes the results of ImagineNet-FC and its variants based on TSN, TSM, and ST-GCN.
Performance of the random weighted summation mechanism surpasses the vanilla method and other two bilinear pooling aggregation methods both in RGB and pose modes.
This reveals that the proposed mechanism can generate richer feature combinations concisely and effectively, thus enabling ImagineNet to achieve better generalization performance on unseen error combinations.

\begin{table}[t] \footnotesize \renewcommand \arraystretch{0.58}
    \caption{Ablation studies on three feature aggregation strategies.}
    \label{Tab7-Ablation}
    \vspace{-15pt}
\begin{center}
    \begin{tabular}{cccccc}
        \toprule
        \tabcolsep=1cm
        Model & Agg-1 & Agg-2 & Agg-3 & mAP & mmit mAP \\
        \midrule
        TSN \cite{TSN2016ECCV} & -- & -- & -- & 0.5598 & 0.6143 \\
        \midrule
        \multirow{4}*{\textit{w}/ ImagineNet-FC} & \ding{56} & \ding{56} & \ding{56} & \underline{0.6198} & 0.6738 \\
        ~ & \ding{52} & \ding{56} & \ding{56} & \textbf{0.6259} & \textbf{0.6893} \\
        ~ & \ding{56} & \ding{52} & \ding{56} & 0.6019 & \underline{0.6775} \\
        ~ & \ding{56} & \ding{56} & \ding{52} & 0.6033 & 0.6725 \\
        \midrule \midrule
        TSM \cite{TSM2019ICCV} & -- & -- & -- & 0.5662 & 0.6618 \\
        \midrule
        \multirow{4}*{\textit{w}/ ImagineNet-FC} & \ding{56} & \ding{56} & \ding{56} & \underline{0.6871} & \underline{0.7353} \\
        ~ & \ding{52} & \ding{56} & \ding{56} & \textbf{0.7053} & \textbf{0.7566} \\
        ~ & \ding{56} & \ding{52} & \ding{56} & 0.6434 & 0.7308 \\
        ~ & \ding{56} & \ding{56} & \ding{52} & 0.6569 & 0.7219 \\
        \midrule \midrule
        ST-GCN \cite{ST-GCN} & -- & -- & -- & 0.5776 & 0.6692 \\
        \midrule
        \multirow{4}*{\textit{w}/ ImagineNet-FC} & \ding{56} & \ding{56} & \ding{56} & \underline{0.6374} & \underline{0.7089} \\
        ~ & \ding{52} & \ding{56} & \ding{56} & \textbf{0.6404} & \textbf{0.7115} \\
        ~ & \ding{56} & \ding{52} & \ding{56} & 0.5783 & 0.6877 \\
        ~ & \ding{56} & \ding{56} & \ding{52} & 0.6159 & 0.6864 \\
        \bottomrule
    \end{tabular}
\end{center}
\end{table}

\subsection{Cross Modality Studies}
In previous experiment settings, inputs of the ImaginNet belong to different categories but the same modality.
The structure of ImagineNet inherently supports multi-modal data fusion.
Taking TSN, TSM and ST-GCN as basic models, Table \ref{TSN-Cross-modality} and Table \ref{TSM-Cross-modality} compares the ImagineNet-CA with the Two-Stream fusion method and two bilinear pooling fusion methods under cross modality settings. 
The latency of these fusion models is reported by averaging 1000 running times, while basic models are excluded. 
Results show that the ImagineNet-CA surpasses the other three multimodal fusion methods.
Although BLOCK performs similarly to ImagineNet-CA, its latency is nearly 7.8$\times$ longer, which is mainly caused by the complex approximate outer product computation.
The Two-Stream fusion model can reduce latency but has poor performance.

\section{Limitation and Discussion}

As the first study on fine-grained error action recognition and AQA in CPR training, this work inevitably has some limitations.
The diversity and complexity of the CPR-Coach dataset remains to be improved. 
Standard CPR \cite{AHA_Guidelines} consists of several stages (\textit{e.g.}, electric defibrillation, artificial respiration), while only the external cardiac compression is studied due to the time and scale limitation.
Nevertheless, the CPR-Coach has reached 449GB and 2.2M frames.
In the future, we will continue to cooperate with the training center of the hospital to enrich the CPR-Coach dataset.
There is still huge potential exploration space for complex and multi-stage medical action analysis.

\begin{table}[t] \footnotesize \tabcolsep=2pt \renewcommand \arraystretch{0.7}
    \caption{Cross modality studies on \textit{RGB} and \textit{Pose} information based on TSN and ST-GCN.}
    \label{TSN-Cross-modality}
    \vspace{-15pt}
\begin{center}
    \begin{tabular}{ccccc}
        \toprule
        Model & Modality & Latency (ms)$\downarrow$ & mAP & mmit mAP \\
        \midrule
        TSN \cite{TSN2016ECCV} & RGB & --- & 0.5598 & 0.6143 \\
        ST-GCN \cite{ST-GCN} & Pose & --- & 0.5776 & 0.6692 \\
        \midrule
        Two-Stream \cite{Two-Stream} & RGB+Pose & \textbf{0.1426} & 0.5915 & 0.6823 \\
        CBP \cite{CBP} & RGB+Pose & 0.3032 & 0.7066 & 0.7460 \\
        BLOCK \cite{Block} & RGB+Pose & 1.254 & \underline{0.7094} & \underline{0.7597} \\
        \midrule
        \textit{w}/ ImagineNet-CA & RGB+Pose & \underline{0.1612} & \textbf{0.7133} & \textbf{0.7641} \\
        \bottomrule
    \end{tabular}
\end{center}
\end{table}
\begin{table}[t] \footnotesize \tabcolsep=2pt \renewcommand \arraystretch{0.7}
    \caption{Cross modality studies on \textit{RGB} and \textit{Pose} information based on TSM and ST-GCN.}
    \label{TSM-Cross-modality}
    \vspace{-15pt}
\begin{center}
    \begin{tabular}{ccccc}
        \toprule
        Model & Modality & Latency (ms)$\downarrow$ & mAP & mmit mAP \\
        \midrule
        TSM \cite{TSM2019ICCV} & RGB & --- & 0.5662 & 0.6618 \\
        ST-GCN \cite{ST-GCN} & Pose & --- & 0.5776 & 0.6692 \\
        \midrule
        Two-Stream \cite{Two-Stream} & RGB+Pose & \textbf{0.1501} & 0.6003 & 0.6815 \\
        CBP \cite{CBP} & RGB+Pose & 0.3043 & 0.7089 & 0.7506 \\
        BLOCK \cite{Block} & RGB+Pose & 1.294 & \underline{0.7107} & \textbf{0.7675} \\
        \midrule
        \textit{w}/ ImagineNet-CA & RGB+Pose & \underline{0.1642} & \textbf{0.7110} & \underline{0.7515} \\
        \bottomrule
    \end{tabular}
\end{center}
\end{table}

\section{Conclusion}
This paper proposes the CPR-Coach dataset, which can support fine-grained action recognition and composite error action recognition tasks under restricted supervision.
In the first task, we extensively evaluate and compare the existing HAR models. 
In the second task, we propose different ImagineNet frameworks inspired by human cognition to improve the performance of the model under the composite error settings.
Sufficient experiments verified the effectiveness of the framework.
We hope this work can bring new inspiration and actual contribution to the computer vision and medical skills training communities simultaneously.


\section*{Acknowledgments}
This work was supported in part by the National Key R\&D Program of China (2021ZD0113502); the Shanghai Municipal Science and Technology Major Project (2021SHZDZX0103).

\bibliography{mybibfile}

\end{document}